\documentclass[letterpaper]{article}
\usepackage[preprint]{aaai2027}
\usepackage[hyphens]{url}
\usepackage{graphicx}
\usepackage{natbib}
\usepackage{caption}
\usepackage{booktabs}
\usepackage{tabularx}
\usepackage{amsmath}
\usepackage{amssymb}
\usepackage{multirow}
\usepackage{colortbl}
\usepackage{placeins}
\usepackage{afterpage}
\usepackage{booktabs}
\usepackage{multirow}
\usepackage{colortbl}
\usepackage{amsmath}
\usepackage{amsfonts}
\usepackage{amssymb}
\usepackage{array}
\usepackage{tabularx}
\usepackage{pifont}
\usepackage{enumitem}
\usepackage[most,skins,theorems]{tcolorbox}
\usepackage{algorithm}
\usepackage{algpseudocode}
\newcommand{\algcomment}[1]{\Comment{#1}}
\usepackage{xspace}

\definecolor{codelinkblue}{RGB}{30,144,255}

\tcbset{
  aibox/.style={
    width=\linewidth,
    top=8pt,
    bottom=4pt,
    colback=white,
    colframe=black,
    colbacktitle=black,
    enhanced,
    center,
    attach boxed title to top left={yshift=-0.1in,xshift=0.15in},
    boxed title style={boxrule=0pt,colframe=white},
  }
}
\newtcolorbox{AIbox}[2][]{aibox,title={#2},#1}

\newcommand{\method}{GPD\xspace}

\title{Distilling Routed 3D Privilege for Spatial Reasoning in Vision-Language Models}

\author{
Hongxing Li\textsuperscript{\rm 1,}\equalcontrib,
Yixin Li\textsuperscript{\rm 1,}\equalcontrib,
Dingming Li\textsuperscript{\rm 1},
Zixuan Wang\textsuperscript{\rm 1},
Yuchen Yan\textsuperscript{\rm 1},\\
Wenqi Zhang\textsuperscript{\rm 1},
Weiming Lu\textsuperscript{\rm 1},
Yongliang Shen\textsuperscript{\rm 1,}\corresponding
}

\affiliations{
\textsuperscript{\rm 1}Zhejiang University \\

\texttt{\{hongxing.li, syl\}@zju.edu.cn}
}

\begin{document}

\maketitle

\pagenumbering{arabic}
\pagestyle{plain}
\thispagestyle{plain}

\afterpage{
\begin{figure*}[h]
\centering
\includegraphics[width=0.9\textwidth]{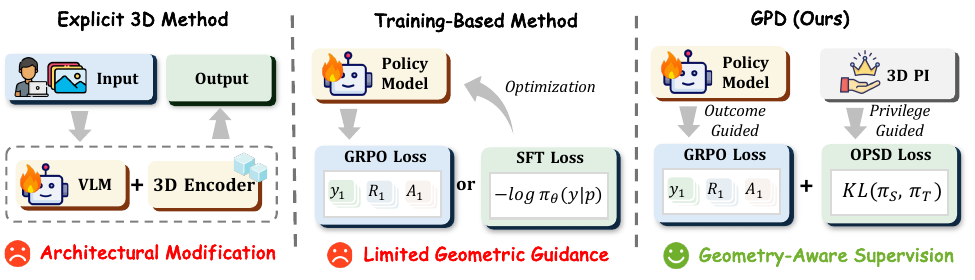}
\caption{
Comparison of three paradigms for spatial reasoning in VLMs: explicit 3D methods that require architectural modification, training-based methods that rely on outcome-only supervision, and GPD that introduces question-routed 3D cues as training-time privilege.
}
\label{fig:motivation}
\end{figure*}
}

\begin{abstract}

Spatial reasoning remains a persistent weakness of vision-language models (VLMs), because RGB inputs do not directly provide geometric evidence. Existing remedies either inject 3D into the model at inference, paying architecture and latency costs, or train with outcome rewards that supervise only the final answer. Spatial errors originate in perception: a misjudged depth or direction can be corrected only by the scene's true geometry, which the 3D-scanned sources of spatial training corpora already provide. We propose \textbf{\method} (\textbf{G}eometry-\textbf{P}rivileged \textbf{D}istillation), which makes geometric evidence the privilege in on-policy self-distillation (OPSD). For each question, depth, semantic, and bird's-eye-view (BEV) cues are rendered as compact text and routed to the teacher alongside the reference answer; a privileged KL, applied only to incorrect trajectories, augments GRPO, and the deployed model remains RGB-only.
On the 4B backbone, \method achieves 57.1 on VSI-Bench and 37.6 average across MindCube, SPAR-Bench, MMSI-Bench, and ViewSpatial, outperforming both GRPO and answer-privileged OPSD across spatial reasoning benchmarks. Ablations confirm the complementarity of 3D and answer privilege, the advantage of question-conditioned routing over full-context injection, and the benefit of restricting distillation to incorrect trajectories. Our code is available at \textcolor{codelinkblue}{\texttt{https://github.com/ZJU-REAL/GPD}}.

\end{abstract}

\section{Introduction}

Spatial reasoning requires a vision-language model (VLM) to judge distance, direction, and layout, as in planning a route, grasping the nearest object, or estimating how far a sofa is from a door. Benchmarks consistently show that these judgments remain difficult for current VLMs~\cite{yang2025vsibench,wang2025mindcube,zhang2025spar,yang2025mmsi,li2025viewspatial}. The difficulty stems from the input: RGB frames do not directly provide geometric evidence, so the model must infer depth and layout from appearance alone.

Two lines of work address this gap. Explicit 3D methods integrate geometric encoders or representations into the model~\cite{chen2024spatialvlm,cheng2024spatialrgpt,wu2026spatial-mllm,zheng2025video3d-llm}, or invoke perception and reconstruction tools at test time~\cite{han2025tiger,chen2026gca,cho2026spatialclaw,qi2025gpt4scene}, at the cost of architectural changes and additional inference overhead. Training-based methods adapt the model on spatial question-answering data through supervised fine-tuning or outcome-based reinforcement learning~\cite{ouyang2025spacer,li2025spatialladder,ma2025spatialreasoner}; RGB-only inference is preserved, but supervision comes only from the final answer, so a geometric error along the reasoning chain, such as a wrong depth order or a flipped left--right relation, is neither localized nor corrected.

The two lines thus sit on opposite sides of a trade-off: explicit 3D methods gain geometric evidence but add inference overhead, while training-based methods keep inference RGB-only but train without geometric supervision. 
Yet at training time, geometry is readily available: spatial corpora 
derived from 3D scans provide depth, semantics, and layout directly, 
and off-the-shelf models can estimate the same signals otherwise.
 \textbf{These 3D signals cannot serve as inputs at deployment, but they can supervise training as privileged information.} The question is how to turn this privilege into supervision that an RGB-only model can internalize.

A natural starting point is on-policy self-distillation (OPSD)~\cite{zhao2026opsd,yang2026rlsd,xu2026passbatontrajectoryrelayedonpolicy}, which converts privilege into supervision: the student samples responses from the task input alone, while the same model, additionally given the privileged context, scores each sampled token as a teacher. In spatial reasoning, however, errors originate in perception: the model misjudges depth, position, or extent, then reasons coherently over the wrong facts. The reference answer, the standard privilege in OPSD, cannot correct such errors: \textbf{a misjudged depth or direction can be corrected only by the scene's true geometry}. The privilege must instead supply geometric evidence matched to the question: counting, distance, and layout each rest on different evidence.

We introduce \textbf{\method}, an on-policy self-distillation framework that provides the teacher with question-relevant geometric evidence. For each query, \method constructs textual cues from depth, semantic segmentation, and bird's-eye-view (BEV) layouts, and presents the routed cues to the teacher together with the reference answer. The student samples trajectories from RGB observations and the question, while the same VLM, conditioned on the privileged context, re-scores these trajectories token by token. \method combines this distillation signal with GRPO and applies the privileged KL only to incorrect trajectories, concentrating process supervision on trajectories that require correction. The privilege thus enters only through the training signal, and the deployed model remains RGB-only. Figure~\ref{fig:motivation} illustrates the distinction between \method and the two prior paradigms.

We evaluate \method with Qwen3-VL-Instruct backbones at 2B and 4B scales on five spatial reasoning benchmarks. On the 4B backbone, \method achieves 57.1 on VSI-Bench~\cite{yang2025vsibench} and 37.6 average across MindCube~\cite{wang2025mindcube}, SPAR~\cite{zhang2025spar}, MMSI~\cite{yang2025mmsi}, and ViewSpatial~\cite{li2025viewspatial}, outperforming both GRPO and answer-privileged OPSD. Ablations confirm the effectiveness of each component. Further analysis indicates that the geometric cues have been internalized by the student model. Together, these results establish question-routed geometric privilege as effective supervision for spatial reasoning.

Our contributions are threefold:
\begin{itemize}
\item We formulate geometry-aware privileged self-distillation for spatial reasoning, where the teacher is equipped with spatially grounded evidence, enabling meaningful token-level distillation along incorrect reasoning trajectories.
\item We propose \method, equipping an OPSD teacher with question-routed depth, semantic, and BEV cues, concentrating token-level distillation on incorrect trajectories while retaining RGB-only inference.
\item Experiments across two model scales and five benchmarks demonstrate consistent improvements over GRPO and answer-privileged OPSD, with ablations validating the key design choices.
\end{itemize}

\section{Related Work}

\paragraph{Spatial Reasoning in VLMs.}
Existing work improves spatial reasoning in VLMs through explicit 3D augmentation or training-based adaptation. Explicit methods integrate geometric encoders or representations into VLMs~\cite{chen2024spatialvlm,cheng2024spatialrgpt,wu2026spatial-mllm}, or use auxiliary perception, reconstruction, and tool-based pipelines at inference time~\cite{han2025tiger,chen2026gca,cho2026spatialclaw}. Training-based methods instead apply supervised fine-tuning or outcome-based reinforcement learning to spatial question-answering data~\cite{ouyang2025spacer,li2025spatialladder,ma2025spatialreasoner}. \method connects these directions by using 3D cues as teacher-side training privilege, providing geometry-aware token-level supervision while leaving the student architecture and inference inputs unchanged.

\paragraph{On-policy Self-distillation.}
Knowledge distillation transfers information through teacher distributions~\cite{hinton2015distilling}. OPSD extends this idea to on-policy reasoning by conditioning a teacher on privileged information and re-scoring student rollouts token by token~\cite{zhao2026opsd,yang2026rlsd}. Recent studies examine the stability and refinement of this supervision~\cite{kim2026opsdcompress,shen2026purifiedopsd,kaur2026rethinkingopsd}, while multimodal variants use privileged crops, zoomed views, or visual-thought traces~\cite{yuan2026visionopd,cai2026imagineopd,li2026visualopsd}. \method instead constructs question-routed textual privilege from depth, semantics, and BEV layouts, and applies distillation only to incorrect trajectories.

\begin{figure*}[t]
\centering
\includegraphics[width=\textwidth]{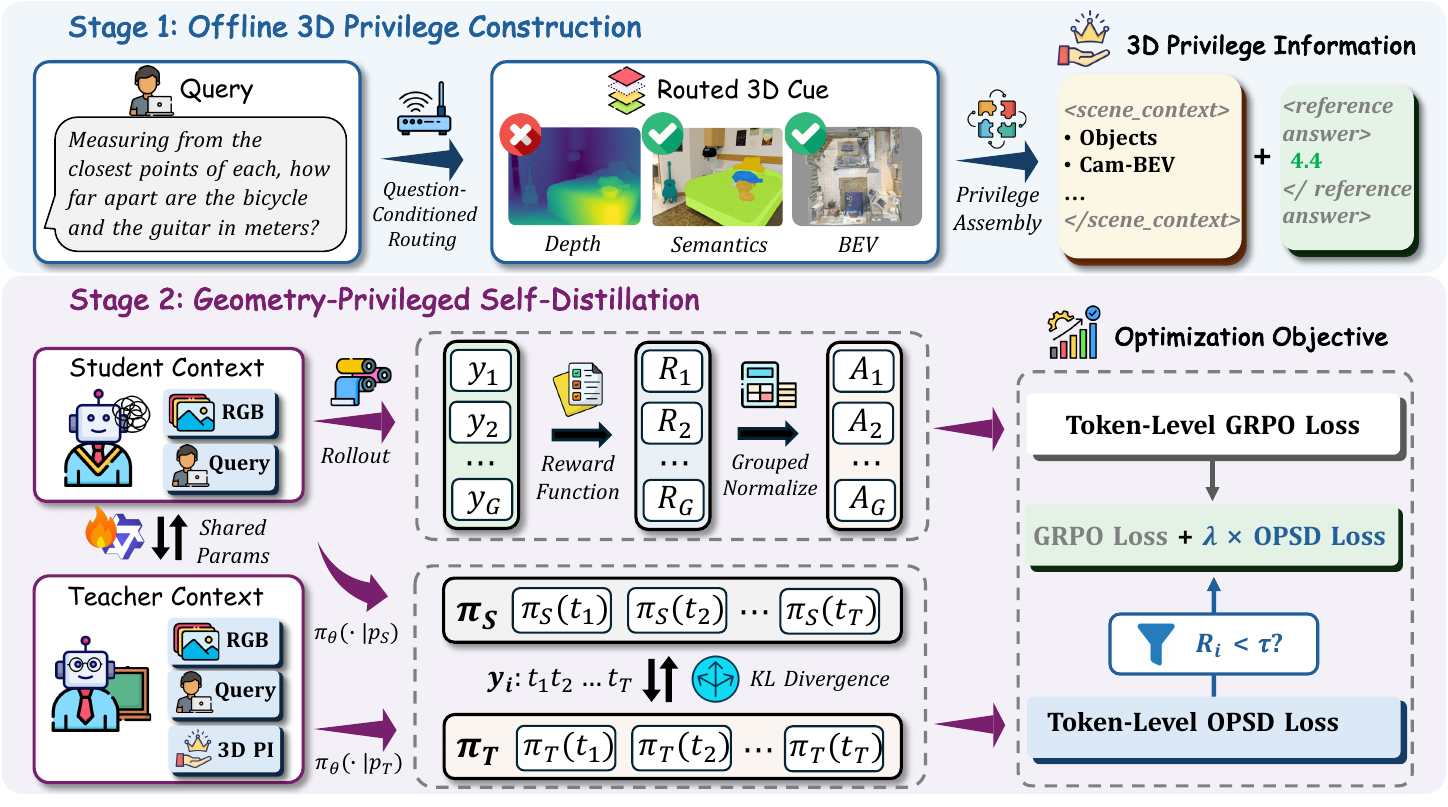}
\caption{
Overview of \method. Stage 1 constructs 3D privilege offline and routes 
question-relevant cues to form the teacher context. Stage 2 trains the same 
VLM as student and teacher: the student samples on-policy trajectories optimized 
by GRPO; the teacher provides token-level KL supervision selectively to incorrect 
trajectories as the OPSD loss. The two losses are jointly optimized.
}
\label{fig:method}
\end{figure*}

\section{Method}

\subsection{Overview of \method}

\method addresses a core tension in spatial reasoning: 3D geometric cues such as depth, layout, and object positions are highly informative for training, yet unavailable at inference when only RGB images and text are observed. Rather than discarding this information or requiring it at test time, \method uses it as training-time privilege to guide policy learning.

As illustrated in Figure~\ref{fig:method}, the method proceeds in two stages. First, for each training sample, \method constructs compact textual summaries of depth, semantic, and BEV cues offline, then routes only the question-relevant subset to form a privileged context for the teacher. Second, during training, the same VLM plays two roles: as a student conditioned only on RGB and question, it samples on-policy trajectories; as a privileged teacher additionally conditioned on the routed 3D cues and reference answer, it re-scores those trajectories token by token. Training combines outcome-based GRPO~\cite{shao2024deepseekmath} with privilege-guided distillation applied selectively to incorrect trajectories, leaving the inference interface unchanged.

\subsection{Offline 3D Privilege Construction}

We represent each training example as $(v,q,a^\star)$, where $v$ denotes the RGB observations, $q$ the question, and $a^\star$ the answer. For each example, \method{} constructs an available subset of three 3D cues offline: depth, semantic, and BEV.

\paragraph{Textual 3D Cue Extraction.}
When aligned 3D annotations are available, we derive depth, semantic, 
and BEV cues directly from them; otherwise, we estimate depth via 
Depth Anything~3~\citep{lin2025da3} and obtain semantic instance masks 
and labels via Grounded-SAM-2~\citep{ren2024sam}, with BEV cues 
included only when reliable camera poses and scene-layout information 
are available. All cues are computed once and fixed throughout 
training. Let $\mathcal{M}\subseteq\{\text{depth},\text{semantic},
\text{BEV}\}$ denote the modalities available for a given example. 
For each $m\in\mathcal{M}$, a fixed template converts its cue into a 
textual summary $s_m$: depth records object distances and near--far 
order, semantic lists objects with their image regions, and BEV 
describes top-down relations. Frame identifiers are 
retained for multi-frame inputs.

\paragraph{Question-Conditioned Routing.}
Different spatial questions require different forms of geometric evidence: counting questions rely on semantic and visibility cues, whereas distance and navigation questions benefit from depth and BEV. Injecting all modalities indiscriminately may introduce irrelevant context and dilute the teacher signal. GPD therefore performs question-conditioned routing offline, using the base VLM to select a relevant subset $r(q) \subseteq \mathcal{M}$ for each question. The routed scene context is

\begin{equation}
c_{\mathrm{scene}}(q)=\operatorname{concat}(s_m:m\in r(q)),
\end{equation}
and the complete privilege provided to the teacher is
\begin{equation}
c_{\mathrm{priv}}(q,a^\star)=\operatorname{concat}(c_{\mathrm{scene}}(q),a^\star).
\end{equation}

We present an example of the teacher privilege as follows.

\begin{AIbox}{Example of Teacher Privilege}
\small
\textit{
\textbf{<scene\_context>}\\
\textbf{Room layout:}\\
\footnotesize{
chair(west), table(center), sofa(northwest), door(east), wall(north)
}\\
\textbf{[frame\_00]}\\
\textbf{Depth near$\rightarrow$far:}\\
\footnotesize{
chair(1.39m)$\rightarrow$table(2.08m)$\rightarrow$sofa(2.64m)$\rightarrow$door(3.05m)
}\\
\textbf{Objects:}\\
\footnotesize{
chair [right-bottom, 5\%], table [center, 11\%], sofa [left-middle, 18\%], door [right-middle, 3\%]
}\\
\textbf{Cam-BEV:}\\
\footnotesize{
chair(left-front, 1.4m), table(front, 2.1m), sofa(left-front, 2.6m), door(right-front, 3.1m)
}\\
\textbf{</scene\_context>}\\
\\
\textbf{<reference\_answer>}\\
\textbf{A}\\
\textbf{</reference\_answer>}
}
\end{AIbox}

\begin{figure*}[t]
\centering
\includegraphics[width=\textwidth]{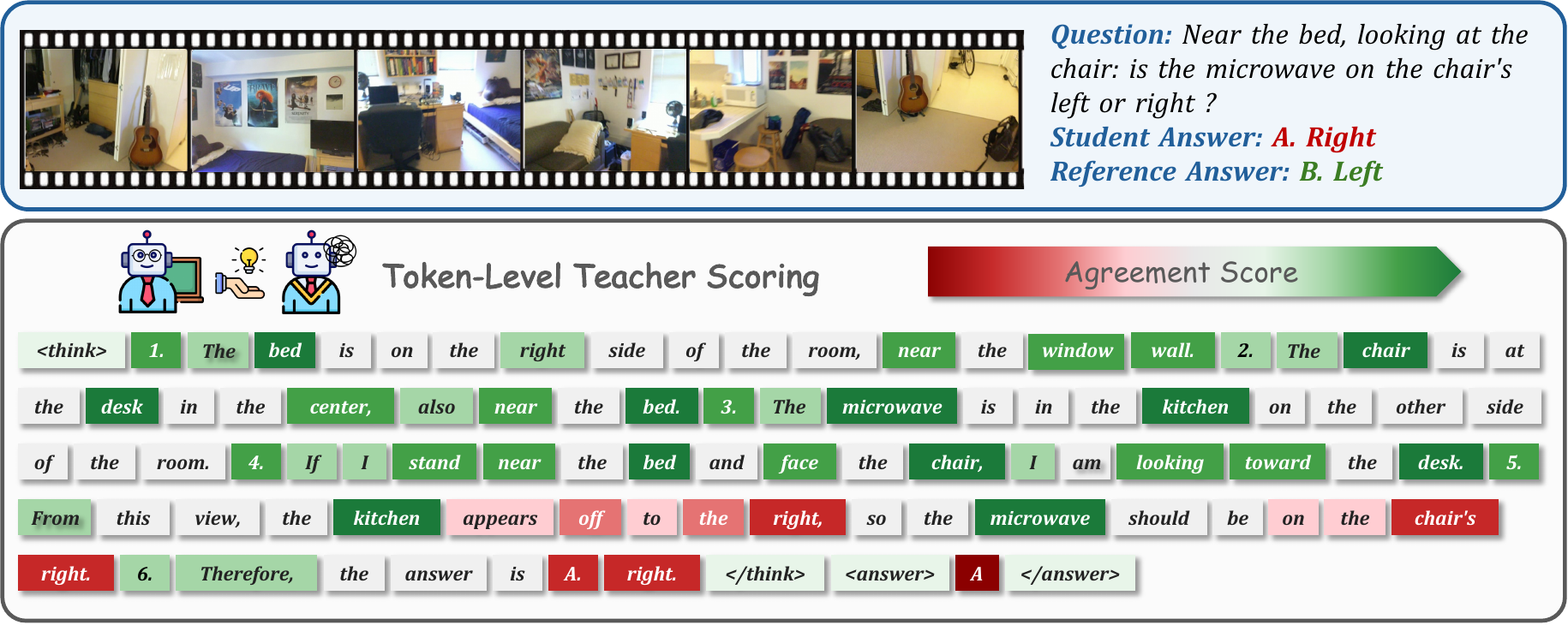}
\caption{
Token-level teacher scoring on a student trajectory. 
The student answers incorrectly (A. Right) while the reference answer is B. Left. 
The teacher assigns high agreement scores to spatially grounded tokens 
and progressively rejects tokens where the reasoning goes astray.
}
\label{fig:scoring}
\end{figure*}

\subsection{Geometry-Privileged Self-Distillation}

During training, the same VLM serves as both student and teacher under different conditioning contexts. The student receives only the RGB observations and question, $p_S=\operatorname{concat}(v,q)$, while the teacher additionally receives the offline-constructed privilege, $p_T=\operatorname{concat}(v,q,c_{\mathrm{priv}}(q,a^\star))$.

\paragraph{Outcome-Guided Policy Optimization.}
For each training example, the student samples a group of $G$ responses $\{y_i\}_{i=1}^{G}\sim\pi_\theta(\cdot\mid p_S)$ using only the student context $p_S$, where each response $y_i=(t_1,t_2,\ldots,t_T)$ consists of $T$ tokens. A group-normalized advantage $A_i$ is computed from the outcome reward $R_i$ of $y_i$. We then optimize the following GRPO objective over all sampled responses
\begin{equation}
\begin{aligned}
\mathcal{L}_{\mathrm{GRPO}}(\theta)
=
-\mathbb{E}_{\{y_i\}_{i=1}^{G}}
\Bigg[
\frac{1}{G}\sum_{i=1}^{G}
\frac{1}{T}\sum_{j=1}^{T}
\min\!\Big(
\rho_{i,j} A_i,\,
\\
\operatorname{clip}(\rho_{i,j},1\pm\varepsilon)A_i
\Big)
-\beta\,\frac{1}{G}\sum_{i=1}^{G}D_{\mathrm{KL}}\!\left[\pi_\theta\,\|\,\pi_{\mathrm{ref}}\right]_i
\Bigg],
\end{aligned}
\end{equation}
where $\rho_{i,j}=\pi_\theta(t_j\mid t_{<j},p_S)\,/\,\pi_{\theta_{\mathrm{old}}}(t_j\mid t_{<j},p_S)$ is the token-level importance sampling ratio, and $D_{\mathrm{KL}}[\pi_\theta\|\pi_{\mathrm{ref}}]=\sum_{j=1}^{T}\left[\log\pi_\theta(t_j\mid t_{<j},p_S)-\log\pi_{\mathrm{ref}}(t_j\mid t_{<j},p_S)\right]$ is the token-level KL divergence with the frozen reference policy $\pi_{\mathrm{ref}}$. Here, $\beta$ controls the strength of regularization toward the reference policy.

\paragraph{Privilege-Guided Policy Distillation.}
At each token position $j$, the teacher and student induce conditional distributions 
$\pi_\theta(\cdot\mid t_{<j}, p_T)$ and $\pi_\theta(\cdot\mid t_{<j}, p_S)$ respectively. 
The per-token reverse KL divergence is expensive to compute due to the full-vocabulary 
summation. We therefore take a single-sample estimate on the student-sampled token 
$t_j\sim\pi_\theta(\cdot\mid t_{<j},p_S)$, yielding the token-level teacher--student 
log-probability gap
\begin{equation}
\Delta_{i,j} = \log\pi_\theta(t_j\mid t_{<j},p_T)-\log\pi_\theta(t_j\mid t_{<j},p_S),
\end{equation}
and its corresponding reverse KL estimate $d_{i,j} = \exp(\Delta_{i,j}) - \Delta_{i,j} - 1$.
The distillation loss is then aggregated over all valid tokens and responses as
\begin{equation}
\mathcal{L}_{\mathrm{OPSD}}(\theta) = 
\sum_{i=1}^{G}\sum_{j=1}^{T} u_{i,j}\, d_{i,j} \Big/ \sum_{i=1}^{G}\sum_{j=1}^{T} u_{i,j},
\end{equation}
where $u_{i,j}\in\{0,1\}$ indicates a valid response token. This aligned comparison provides token-level supervision along the student's own reasoning trajectory, rather than on teacher-generated rollouts. Figure~\ref{fig:scoring} illustrates the token-level teacher scoring on a representative incorrect trajectory.

\paragraph{Error-Gated Joint Objective.}
Directly applying distillation on all trajectories is suboptimal: when the student already 
produces correct responses, the teacher signal may introduce noise or conflicting gradients. 
We therefore gate $\mathcal{L}_{\mathrm{OPSD}}$ to incorrect trajectories only, leaving correct ones unaffected. To further suppress occasionally large distillation signals, we aggregate the OPSD loss over valid tokens
from incorrect trajectories within each optimization micro-batch and clip the
resulting scalar at $1.5$ before scaling:
\begin{equation}
\tilde{\mathcal{L}}_{\mathrm{OPSD}} = \min\!\left(
\sum_{i\in\mathcal{E}}\sum_{j=1}^{T} u_{i,j}\, d_{i,j} \Big/ \sum_{i\in\mathcal{E}}\sum_{j=1}^{T} u_{i,j},\;1.5
\right),
\end{equation}
where $\mathcal{E}=\{i:R_i<\tau\}$ denotes the set of incorrect trajectories and we set
$\tau=0.5$. The final joint objective is
\begin{equation}
\mathcal{L} = \mathcal{L}_{\mathrm{GRPO}} + \lambda\,\tilde{\mathcal{L}}_{\mathrm{OPSD}},
\end{equation}
where $\lambda$ controls the strength of the distillation signal.

\begin{table*}[t!]
\centering
\small
{\setlength{\tabcolsep}{1mm}%
\begin{tabular*}{\textwidth}{@{\extracolsep{\fill}}lccccccccc}
\toprule
\multirow{2}{*}{\textbf{Model}} &
\multirow{2}{*}{\textbf{Avg.}} &
\multicolumn{4}{c}{\textbf{Numerical Question}} &
\multicolumn{4}{c}{\textbf{Multiple-choice Question}} \\
\cmidrule(lr){3-6}\cmidrule(lr){7-10}
& & Obj. Cnt & Abs. Dist. & Obj. Size & Room Size
& Rel. Dist. & Rel. Dir. & Route Plan. & Appr. Order \\
\midrule
\rowcolor{gray!10}
\multicolumn{10}{@{}l}{\textit{\textbf{General-Purpose VLMs}}} \\
GPT-4o~\citep{hurst2024gpt} & 34.0 & 46.2 & 5.3 & 43.8 & 38.2 & 37.0 & 41.3 & 31.5 & 28.5 \\
Gemini-3-Pro~\citep{team2024gemini} & 52.5 & 38.0 & 37.8 & 72.7 & 44.1 & 59.9 & 55.7 & 45.9 & 66.0 \\
LLaVA-OneVision-72B~\citep{li2024llava-one} & 40.3 & 43.5 & 23.9 & 57.6 & 37.5 & 42.5 & 39.9 & 32.5 & 44.6 \\
Qwen2.5-VL-7B~\citep{bai2025qwen25vl} & 35.8 & 43.5 & 15.1 & 48.5 & 41.1 & 36.3 & 40.1 & 28.4 & 33.7 \\
\midrule
\rowcolor{gray!10}
\multicolumn{10}{@{}l}{\textit{\textbf{3D-Augmented Methods}}} \\
Spatial-MLLM-4B~\citep{wu2026spatial-mllm} & 47.3 & 65.6 & 35.5 & 64.2 & 40.6 & 41.3 & 47.9 & 34.0 & 49.2 \\
Think3D-4B~\citep{zhang2026think3d} & - & - & - & - & - & 44.7 & 39.0 & 36.7 & 61.2 \\
\midrule
\rowcolor{gray!10}
\multicolumn{10}{@{}l}{\textit{\textbf{Outcome-Supervised Methods}}} \\
SpaceR-7B~\citep{ouyang2025spacer} & 44.5 & 63.2 & 30.0 & 60.3 & 37.6 & 39.7 & 45.6 & 31.4 & 48.2 \\
SpatialLadder-3B~\citep{li2025spatialladder} & 45.7 & 63.5 & 34.3 & 61.7 & 43.9 & 45.4 & 44.8 & 35.6 & 36.4 \\
\midrule
\rowcolor{gray!10}
\multicolumn{10}{@{}l}{\textit{\textbf{Controlled Training Variants}}} \\
\midrule
Qwen3-VL-2B~\citep{Qwen3-VL} & 51.3 & 59.9 & 42.1 & 70.3 & \textbf{54.0} & 51.7 & 43.1 & 27.3 & 61.7 \\
\quad \textit{+ GRPO} & 51.3 & 60.2 & 41.2 & \textbf{71.5} & 47.9 & 50.8 & 44.6 & \textbf{29.9} & \textbf{64.4} \\
\quad \textit{+ OPSD} & 50.8 & \textbf{64.1} & 40.8 & 69.4 & 48.1 & 51.4 & 44.7 & \textbf{29.9} & 57.8 \\
\quad \textit{+ \method} & \textbf{51.7} & 61.2 & \textbf{42.1} & 71.3 & 48.7 & \textbf{51.8} & \textbf{45.0} & 29.4 & 64.2 \\
\midrule
Qwen3-VL-4B~\citep{Qwen3-VL} & 55.4 & 67.5 & 43.0 & 74.3 & 61.3 & 52.7 & 47.9 & 33.5 & 62.9 \\
\quad \textit{+ GRPO} & 56.4 & 70.2 & 44.4 & 74.9 & 61.7 & 51.3 & 49.9 & 33.5 & \textbf{65.5} \\
\quad \textit{+ OPSD} & 55.7 & 68.2 & \textbf{46.0} & 73.8 & 58.1 & 51.8 & 48.6 & \textbf{36.6} & 62.6 \\
\quad \textit{+ \method} & \textbf{57.1} & \textbf{70.6} & 44.8 & \textbf{74.9} & \textbf{62.3} & \textbf{54.2} & \textbf{50.7} & 34.0 & 64.9 \\
\bottomrule
\end{tabular*}
}
\caption{VSI-Bench results across numerical and multiple-choice spatial reasoning subtasks. For controlled training variants, bold indicates the best result within each backbone.}
\label{tab:vsi-main}
\end{table*}

\begin{table}[!ht]
\centering
\small
{\setlength{\tabcolsep}{2pt}%
\begin{tabularx}{\columnwidth}{@{}>{\raggedright\arraybackslash}Xccccc@{}}
\toprule
Method & MindCube & SPAR & MMSI & ViewSpatial & Avg. \\
\midrule
GPT-4o & 38.8 & 36.4 & 30.3 & 32.6 & 34.5 \\
Qwen2.5-VL-7B & 29.3 & 30.2 & 25.9 & 37.9 & 30.8 \\
SpatialLadder-3B & 43.5 & 32.9 & 27.4 & 39.9 & 35.9 \\
\midrule
Qwen3-VL-2B & 31.7 & 31.8 & \textbf{24.3} & 36.6 & 31.1 \\
\quad \textit{+ GRPO} & 35.8 & \textbf{38.5} & 19.8 & 36.7 & 32.7 \\
\quad \textit{+ OPSD} & 31.2 & 30.9 & 21.7 & 36.8 & 30.2 \\
\rowcolor{gray!10}
\quad \textit{+ \method} & \textbf{38.4} & 38.2 & 21.1 & \textbf{36.8} & \textbf{33.6} \\
\midrule
Qwen3-VL-4B & 21.5 & 35.2 & 27.7 & 39.1 & 30.9 \\
\quad \textit{+ GRPO} & 33.0 & 41.4 & 28.7 & 39.6 & 35.7 \\
\quad \textit{+ OPSD} & 21.9 & 36.3 & 27.5 & \textbf{40.4} & 31.5 \\
\rowcolor{gray!10}
\quad \textit{+ \method} & \textbf{38.0} & \textbf{43.4} & \textbf{29.4} & 39.8 & \textbf{37.6} \\
\bottomrule
\end{tabularx}
}
\caption{Results on MindCube, SPAR-Bench, MMSI-Bench, and ViewSpatial. For controlled training variants, bold indicates the best result within each backbone.}
\label{tab:other-main}
\end{table}

\section{Experiments}

\subsection{Experimental Setup}

\paragraph{Baselines and Benchmarks.}
We compare \method with general-purpose VLMs~\cite{hurst2024gpt,team2024gemini,li2024llava-one,bai2025qwen25vl}, 3D-augmented methods~\cite{wu2026spatial-mllm,zhang2026think3d}, and outcome-supervised methods~\cite{ouyang2025spacer,li2025spatialladder}. For controlled comparison, we train GRPO and OPSD on the same data and backbone~\cite{shao2024deepseekmath,zhao2026opsd}. We evaluate on five benchmarks~\cite{yang2025vsibench,wang2025mindcube,zhang2025spar,yang2025mmsi,li2025viewspatial}: VSI-Bench covers eight video-based spatial reasoning subtasks; MindCube tests mental spatial modeling from limited views; SPAR-Bench evaluates 3D object relations and scene geometry; MMSI-Bench assesses multi-image spatial intelligence; and ViewSpatial-Bench evaluates perspective-dependent spatial understanding.

\paragraph{Training Dataset.}
We train on 10K VSI, 4K SPAR, and 1K MindCube training-split samples. For VSI and SPAR, which are derived from ScanNet scenes~\cite{dai2017scannet}, privileged information is generated directly from 3D annotations. For MindCube, which lacks aligned 3D ground truth, we estimate depth with Depth Anything 3~\cite{lin2025depthanything3} and obtain semantic instance cues with Grounded-SAM-2~\cite{ren2024groundedsam}. All methods and ablations share the same training mixture and precomputed privileged information.

\paragraph{Training Details.}
All experiments are based on Qwen3-VL-Instruct backbones at 2B and 4B scales~\cite{Qwen3-VL}. We train for one epoch with a learning rate of $1\times10^{-6}$, a rollout group size of $n=8$, and a privileged-distillation weight of $\lambda=0.003$.

\subsection{Main Results}

\begin{figure*}[!t]
\centering
\includegraphics[width=\textwidth]{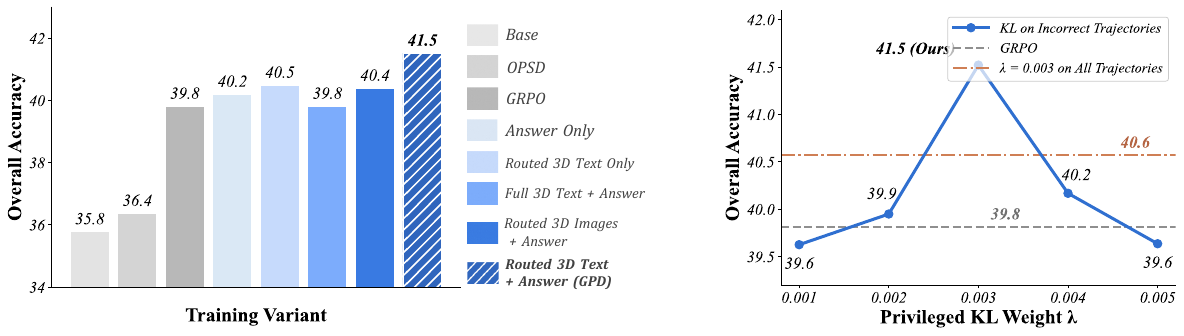}
\caption{Ablation studies on Qwen3-VL-4B. \textbf{Left:} Privilege-component ablation. Grey bars represent no-signal or single-signal baselines; blue bars represent variants that combine outcome reward with privileged KL. \textbf{Right:} Privileged KL weight sweep. The dashed grey line marks GRPO and the dashed orange line marks the all-trajectories variant at $\lambda=0.003$.}
\label{fig:ablation}
\end{figure*}

\paragraph{Results on VSI-Bench.}
Table~\ref{tab:vsi-main} compares \method against general-purpose VLMs, 3D-augmented methods, outcome-supervised methods, and controlled training variants on VSI-Bench. With Qwen3-VL-4B, \method achieves the highest average accuracy of 57.1 among controlled training variants, improving over the base model by 1.7 points, GRPO by 0.7 points, and OPSD by 1.4 points. 
With Qwen3-VL-2B, \method similarly achieves the best controlled average of 51.7, outperforming the base model and GRPO by 0.4 points and OPSD by 0.9 points.
Across both scales, the improvements on geometry-dependent subtasks align with routed 3D privilege providing depth order and spatial layout cues that are difficult to recover from RGB alone.

\paragraph{Results on Additional Benchmarks.}
Table~\ref{tab:other-main} reports results on MindCube, SPAR-Bench, MMSI-Bench, and ViewSpatial. With Qwen3-VL-4B, \method achieves the highest controlled average of 37.6, improving over GRPO by 1.9 points and OPSD by 6.1 points. \method obtains the best result among controlled variants on MindCube (38.0 vs. 33.0 for GRPO, +5.0 points), SPAR (43.4 vs. 41.4 for GRPO, +2.0 points), and MMSI (29.4 vs. 28.7 for GRPO, +0.7 points). With Qwen3-VL-2B, \method again achieves the best controlled average of 33.6, outperforming GRPO by 0.9 points and OPSD by 3.4 points, with the largest gain on MindCube (38.4 vs. 35.8 for GRPO, +2.6 points).

\subsection{Ablation Study}

\paragraph{Effect of Privilege Components.}
Figure~\ref{fig:ablation} (left) ablates the privilege design on Qwen3-VL-4B, reporting Avg.\ as the mean over VSI, MindCube, SPAR, MMSI and ViewSpatial. 
Adding the reference answer alone reaches 40.2 (+0.4 points over GRPO), and adding routed 3D text alone reaches 40.5 (+0.7 points). Combining both yields 41.5, confirming that the two sources are complementary: the answer specifies the target outcome while the 3D context supplies geometric evidence. Full 3D text injection (39.8) falls below routed text by 1.7 points despite containing more information, confirming that indiscriminate injection introduces noise. Routed images reach 40.4, below routed text (41.5) by 1.1 points, suggesting that structured text aligns geometric cues with the symbolic reasoning trajectory better than visual overlays.

\paragraph{Effect of Privileged KL.}
Figure~\ref{fig:ablation} (right) sweeps the privileged KL weight $\lambda$ and its application scope with all other components fixed. Applying privileged KL to all trajectories rather than incorrect ones alone reduces Avg.\ by 0.9 points to 40.6, consistent with noisy teacher signals disturbing already-correct trajectories. Among incorrect-only variants, Avg.\ rises from 39.6 at $\lambda=0.001$ to 41.5 at $\lambda=0.003$ (+1.7 points over GRPO), then decreases to 40.2 at $\lambda=0.004$ and 39.6 at $\lambda=0.005$. Too small a weight provides insufficient process-level correction; too large a weight over-constrains the policy and degrades performance.

\begin{figure*}[!ht]
\centering
\includegraphics[width=\textwidth]{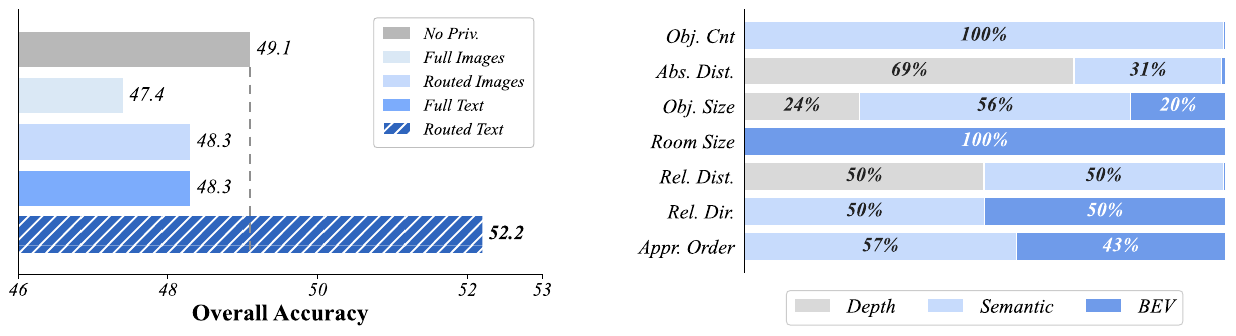}
\caption{Privilege injection analysis on Qwen3-VL-4B-Instruct. \textbf{Left:} Zero-shot overall accuracy under different privilege forms on a VSI-590K subset. \textbf{Right:} Normalized routing weight distribution across VSI task categories and 3D modalities.}
\label{fig:zero_shot_routing}
\end{figure*}

\subsection{Further Analysis}

\paragraph{Zero-Shot Privilege Injection.}
Before RL training, we run a controlled zero-shot study on a $1000$-question VSI-590K subset to decide how 3D privilege should be presented. We evaluate Qwen3-VL-4B-Instruct using identical RGB observations and, when applicable, the same offline question-conditioned routing over depth, semantic segmentation, and BEV, without providing the reference answer. Figure~\ref{fig:zero_shot_routing} (left) summarizes the results. All image-based variants and full text injection remain below the RGB-only baseline ($49.1$): full 3D images drop to $47.4$, while routed images and full text both reach $48.3$. Routed text is the only variant that exceeds the baseline, reaching $52.2$ ($+3.1$ points), motivating its use as the privilege form in \method.

\paragraph{Routing Distribution.}
Figure~\ref{fig:zero_shot_routing} (right) shows the normalized routing distribution across VSI task categories. The pattern aligns well with geometric intuition: object counting is routed entirely to semantic maps, absolute distance relies predominantly on depth, room size is dominated by BEV, and relative direction uses both semantic and BEV. This alignment confirms that the question-conditioned router selects relevant 3D cues rather than injecting every modality.

\begin{figure}[!t]
\centering
\includegraphics[width=\columnwidth]{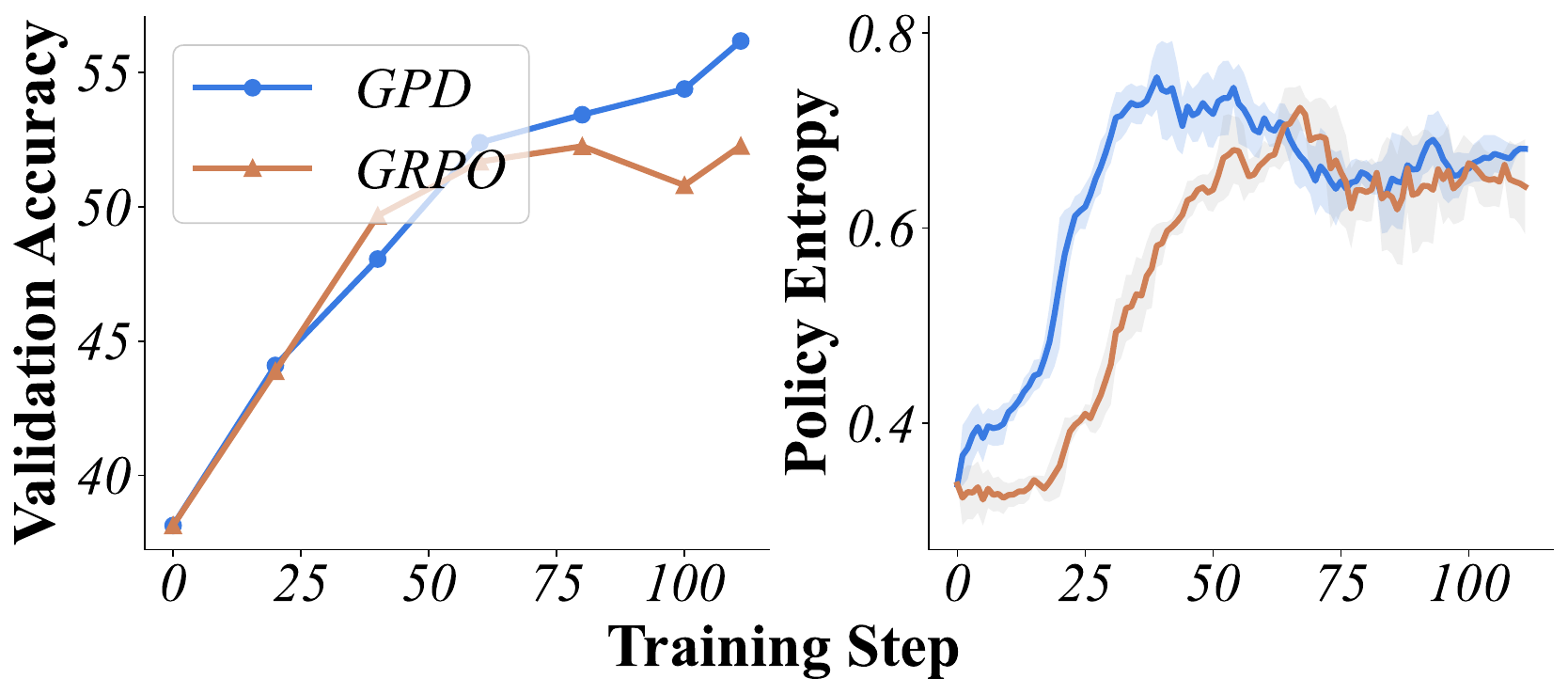}
\caption{Training dynamics of \method and GRPO on Qwen3-VL-4B. Left: validation accuracy. Right: policy entropy.}
\label{fig:training_dynamics}
\end{figure}

\begin{figure}[!t]
\centering
\includegraphics[width=\columnwidth]{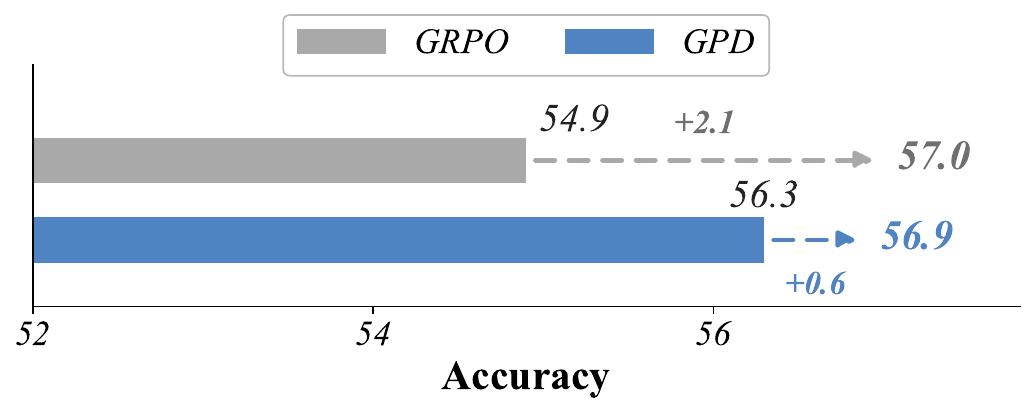}
\caption{Comparison of test-time 3D privilege dependence between GRPO and \method.}
\label{fig:priv-gap}
\end{figure}

\paragraph{Training Dynamics.}
Figure~\ref{fig:training_dynamics} compares \method and GRPO training curves on Qwen3-VL-4B. Both methods show similar validation accuracy in early steps, but \method gradually pulls ahead and continues to improve while GRPO plateaus. The entropy plot reveals why: \method reaches a higher entropy peak earlier than GRPO, indicating that privileged KL encourages broader exploration in the early training phase. As entropy levels align in later steps, \method maintains its accuracy advantage, suggesting that the spatial reasoning modes shaped by privileged supervision are retained.

\paragraph{Privilege Internalization.}
To test whether \method internalizes 3D cues during training, we evaluate models trained on Qwen3-VL-4B with and without privilege injection at test time, as shown in Figure~\ref{fig:priv-gap}. \method ($56.3$) already outperforms GRPO ($54.9$) under RGB-only inference, and gains less when routed 3D text is injected at test time ($+0.6$ vs.\ $+2.1$ for GRPO). The smaller privilege gap suggests that \method internalizes geometric cues during training rather than remaining dependent on them at inference.

\begin{figure}[!t]
\centering
\includegraphics[width=\columnwidth]{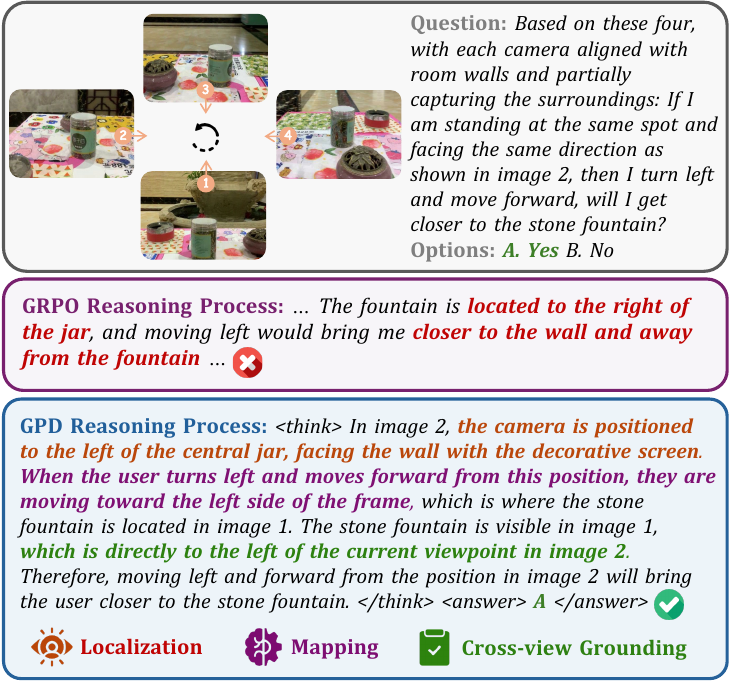}
\caption{Case study comparing \method and GRPO on a multi-view spatial question.}
\label{fig:case_study}
\end{figure}

\paragraph{Case Study.}
Figure~\ref{fig:case_study} shows a representative case where \method answers correctly and GRPO fails. \method correctly localizes the camera, maps the left-forward motion to the spatial frame, and grounds the fountain via cross-view correspondence. GRPO misidentifies the fountain's position relative to the jar and reaches the wrong conclusion. This illustrates a failure mode of outcome-only supervision, where a geometrically incorrect intermediate judgment goes uncorrected in the absence of process-level geometric guidance.

\FloatBarrier

\section{Conclusion}
We present \method, which augments GRPO with an OPSD teacher conditioned on question-routed depth, semantic, and BEV textual cues together with the reference answer. The teacher applies token-level KL supervision selectively to incorrect student trajectories, enabling the student to internalize geometric reasoning during training while retaining RGB-only inference. Across both model scales and benchmarks, \method improves overall accuracy over GRPO and outperforms answer-privileged OPSD.

\bibliography{reference}

\clearpage

\appendix

\twocolumn[
  \begin{center}
    {\LARGE\bfseries Supplementary Material for \\ Distilling Routed 3D Privilege for Spatial Reasoning in Vision-Language Models\par}
    \vspace{0.5em}
  \end{center}
  \vspace{30pt}
]

\section{Methodology Details}

\subsection{Details of Privilege Construction}

\paragraph{3D Cue Construction.}
All 3D cues are constructed offline and used only during training; inference requires no auxiliary model. Depending on the availability of aligned 3D annotations, we employ two source-dependent pipelines.

For VSI and SPAR samples backed by ScanNet~\citep{dai2017scannet}, we project the annotated scene mesh into each camera view with z-buffering to obtain per-frame visible-instance records. Each record stores the semantic label, metric depth, camera-relative horizontal position, image coverage, and a coarse spatial grid location on the image; instances are sorted from near to far. At the scene level, we compute instance centroids on the axis-aligned horizontal plane and describe their positions relative to the room centroid, yielding a bird's-eye-view (BEV) layout cue that captures the global top-down spatial arrangement of the scene and provides a geometric context complementary to the per-frame records.

For MindCube samples that lack aligned annotations, we instead recover pseudo-3D cues directly from RGB observations. Depth Anything~3~\citep{lin2025depthanything3} estimates view-consistent metric depth across the object views, while Grounded-SAM-2~\citep{ren2024groundedsam} produces open-vocabulary instance masks. For each retained instance, we record its semantic label, image position, coverage, and frame-normalized relative depth. Because MindCube lacks reliable camera poses and scene meshes, only depth and semantic cues are made available; BEV is excluded from its modality set to avoid introducing unreliable geometric artifacts. In both pipelines, frames with invalid poses or empty predictions contribute no cue rather than a fabricated fallback, ensuring that the teacher context never introduces spurious geometric information.

\paragraph{3D Cue Routing.}
Routing is performed once before training by the base VLM (Qwen3-VL-4B-Instruct)~\citep{Qwen3-VL}. The router receives only the question text and selects a relevant subset of available 3D modalities, avoiding the injection of geometric context that is irrelevant to the question at hand. For samples not covered by the routing model, we fall back to a rule-based mapping derived from the dataset's question-type annotations to ensure full dataset coverage. When the router selects no modality, no scene context is injected and the teacher context retains only the reference answer, keeping the supervision signal uncontaminated. All routing decisions are stored alongside the precomputed cue records, so the teacher context is assembled deterministically at training time without any additional model inference. The routing prompt is shown below.

\begin{AIbox}{Prompt for 3D Cue Routing}
\small
\textbf{System Prompt}: 

\textit{You are a feature selector for a 3D
spatial reasoning system. Your task is to decide which additional visual
features would help answer a question.}

\medskip

\textbf{User Prompt}: 

\textit{Given the following spatial reasoning
question, decide which additional 3D visual features are needed to answer
it correctly.}

\textit{Available features:}
\begin{itemize}[leftmargin=1.2em,itemsep=1pt,parsep=0pt,topsep=2pt]
  \item \textit{depth -- colorized depth maps showing per-pixel
        distance from the camera; useful for depth prediction, distance
        estimation, and metric values.}
  \item \textit{semantic -- instance segmentation maps where
        each object instance has a unique color; useful for identifying
        named objects, counting, and spatial relations between objects.}
  \item \textit{bev -- bird's-eye view (top-down floor plan) of
        the scene; useful for direction reasoning, viewpoint
        transformation, and floor layout.}
\end{itemize}

\textit{Question: \{question\}}

\textit{Instructions: Output ONLY the feature names that are directly
needed, space-separated. Use exactly these names: depth, semantic, bev.
If the question can be answered from the images alone with no additional
features, output none. Do NOT explain, do NOT output anything else.}

\textit{Examples:}
\begin{itemize}[leftmargin=1.2em,itemsep=1pt,parsep=0pt,topsep=2pt]
  \item \textit{What is the estimated depth of the chair?}
        $\to$ \textit{depth}
  \item \textit{Which is farther, the sofa or the lamp?}
        $\to$ \textit{depth}
  \item \textit{If I face the window, is the door on my left or
        right?} $\to$ \textit{bev}
  \item \textit{How many chairs are visible?} $\to$ \textit{semantic}
  \item \textit{Which object is larger, the table or the cabinet?}
        $\to$ \textit{semantic}
  \item \textit{From the bed, looking toward the closet, where is the
        window?} $\to$ \textit{bev semantic}
\end{itemize}
\end{AIbox}

\subsection{Details of Self-Distillation}

\begin{algorithm}[!t]
\caption{Joint Optimization of GPD}
\label{alg:gpd-training}
\small
\begin{algorithmic}[1]
\Require Policy $\pi_\theta$, frozen reference $\pi_{\mathrm{ref}}$, dataset
$\mathcal{D}$, group size $G$, threshold $\tau$, coefficients $\beta,\lambda$,
valid masks $u_{i,j}$
\For{each optimization micro-batch $\mathcal{B}\subset\mathcal{D}$}
    \State Initialize the GRPO and OPSD accumulators
    \For{each $(v,q,a^\star)$ in $\mathcal{B}$}
        \State $p_S\gets\operatorname{concat}(v,q)$
        \State $p_T\gets\operatorname{concat}(v,q,c_{\mathrm{priv}}(q,a^\star))$

        \algcomment{Step 1: On-policy rollout}
        \State Sample $\{y_i\}_{i=1}^{G}\sim\pi_\theta(\cdot\mid p_S)$
        \State Compute outcome rewards $\{R_i\}_{i=1}^{G}$
        \State $\mu_R\gets\operatorname{mean}(\{R_i\})$,
        $\sigma_R\gets\operatorname{std}(\{R_i\})$
        \State $A_i\gets(R_i-\mu_R)/(\sigma_R+\epsilon)$ for $i=1,\ldots,G$

        \algcomment{Step 2: GRPO loss computation}
        \State $\rho_{i,j}\gets
        \pi_\theta(t_j\mid t_{<j},p_S)/
        \pi_{\theta_{\mathrm{old}}}(t_j\mid t_{<j},p_S)$
        \State $\mathcal{L}_{\mathrm{GRPO}}\gets
        -\operatorname{mean}_{i,j}
        [\min(\rho_{i,j}A_i,
        \operatorname{clip}_{1\pm\varepsilon}(\rho_{i,j})A_i)]$
        \State $\mathcal{L}_{\mathrm{GRPO}}\gets
        \mathcal{L}_{\mathrm{GRPO}}+
        \frac{\beta}{G}\sum_iD_{\mathrm{KL}}[\pi_\theta\|\pi_{\mathrm{ref}}]_i$

        \algcomment{Step 3: OPSD loss computation}
        \For{each valid response position $(i,j)$}
            \State $\ell^T_{i,j}\gets\log\pi_\theta(t_j\mid t_{<j},p_T)$
            \State $\ell^S_{i,j}\gets\log\pi_\theta(t_j\mid t_{<j},p_S)$
            \State $\Delta_{i,j}\gets\ell^T_{i,j}-\ell^S_{i,j}$
            \State $d_{i,j}\gets\exp(\Delta_{i,j})-\Delta_{i,j}-1$
        \EndFor
        \State $\mathcal{L}_{\mathrm{OPSD}}\gets
        \sum_{i,j}u_{i,j}d_{i,j}\big/\sum_{i,j}u_{i,j}$

        \algcomment{Step 4: Error-gated joint optimization}
        \State $\mathcal{E}\gets\{i:R_i<\tau\}$
        \State Accumulate $\{u_{i,j}d_{i,j},u_{i,j}\}_{i\in\mathcal{E}}$
    \EndFor

    \State Set $\tilde{\mathcal{L}}_{\mathrm{OPSD}}\gets 0$ if no token is accumulated
    \State Otherwise, $\tilde{\mathcal{L}}_{\mathrm{OPSD}}\gets
    \min(\sum u_{i,j}d_{i,j}/\sum u_{i,j},1.5)$
    \State $\mathcal{L}\gets\mathcal{L}_{\mathrm{GRPO}}
    +\lambda\tilde{\mathcal{L}}_{\mathrm{OPSD}}$
    \State Update $\theta$ by minimizing $\mathcal{L}$
\EndFor
\end{algorithmic}
\end{algorithm}

\paragraph{GRPO Reward Design.}
For each sampled response $y_i$, we combine a format reward with an
answer-accuracy reward. The format reward encourages a structured response
whose reasoning and final answer are enclosed by
\texttt{<think>}\ldots\texttt{</think>} and
\texttt{<answer>}\ldots\texttt{</answer>}, respectively:
\begin{equation}
r_{\mathrm{fmt}}(y_i)=
\begin{cases}
1, & \text{if $y_i$ contains both blocks},\\
0.5, & \text{if only a valid answer is parsable},\\
0, & \text{otherwise}.
\end{cases}
\end{equation}
Let $\hat{a}_i$ denote the answer parsed from $y_i$, and let $a^\star$
denote the reference answer. For multiple-choice questions, we use exact
matching:
\begin{equation}
r_{\mathrm{mc}}(\hat{a}_i,a^\star)
=\mathbb{I}\!\left[\hat{a}_i=a^\star\right].
\end{equation}
For numerical questions, we use a threshold-averaged relative-accuracy
reward. Let
\begin{equation}
e(\hat{a}_i,a^\star)=
\begin{cases}
\left|\hat{a}_i-a^\star\right|/\left|a^\star\right|, & a^\star\neq 0,\\
\left|\hat{a}_i-a^\star\right|, & a^\star=0,
\end{cases}
\end{equation}
denote the normalized prediction error. The reward averages the success
indicator over a set of confidence thresholds:
\begin{equation}
r_{\mathrm{num}}(\hat{a}_i,a^\star)
=\frac{1}{|\mathcal{T}|}\sum_{\delta\in\mathcal{T}}
\mathbb{I}\!\left[
e(\hat{a}_i,a^\star)\le 1-\delta
\right],
\end{equation}
where $\mathcal{T}=\{0.50,0.55,\ldots,0.95\}$. The accuracy reward is
\begin{equation}
r_{\mathrm{acc}}(\hat{a}_i,a^\star;q)=
\begin{cases}
r_{\mathrm{mc}}(\hat{a}_i,a^\star),
& q\in\mathcal{Q}_{\mathrm{MC}},\\
r_{\mathrm{num}}(\hat{a}_i,a^\star),
& q\in\mathcal{Q}_{\mathrm{Num}},
\end{cases}
\end{equation}
where $\mathcal{Q}_{\mathrm{MC}}$ and $\mathcal{Q}_{\mathrm{Num}}$ denote
the multiple-choice and numerical question sets, respectively. Finally, the
outcome reward used by GRPO is
\begin{equation}
R_i
=0.9\,r_{\mathrm{acc}}(\hat{a}_i,a^\star;q)
+0.1\,r_{\mathrm{fmt}}(y_i).
\end{equation}

\paragraph{Joint Optimization.}

Algorithm~\ref{alg:gpd-training} summarizes the joint optimization procedure
of GPD. Each update first draws on-policy responses from the student context
and computes group-relative advantages from the outcome rewards. The same
response tokens are then re-scored under the privileged teacher context.
Only incorrect trajectories contribute to OPSD, while every trajectory remains
in GRPO.

\paragraph{Student and Teacher Prompts.}
The student and teacher are the same VLM under different conditioning
contexts. The student receives only the RGB frames and the question, and
is instructed to reason inside \texttt{<think>} tags and to give its final
answer inside \texttt{<answer>} tags. The teacher shares the same frames
and question, but its system prompt additionally references the
\texttt{<scene\_context>} and \texttt{<reference\_answer>} blocks and asks
the model to first understand the reference answer and then reason from the
frames and routed 3D cues to derive the answer on its own. The two prompts
are shown below; for numerical questions, the phrase ``final answer'' is
replaced by ``final numeric answer'', and all other wording is unchanged.

\begin{AIbox}{Prompt for the Student}
\small
\textbf{System Prompt}: 

\textit{You are a spatial reasoning assistant.
Please answer based on the video frames.}

\medskip

\textbf{User Prompt}: 

\textit{\{video frames\}}

\textit{Question: \{question\}}

\textit{Think step by step and provide your reasoning inside <think> </think> tags. Then
give your final answer inside <answer> </answer> tags.}
\end{AIbox}

\begin{AIbox}{Prompt for the Teacher}
\small
\textbf{System Prompt}: 

\textit{You are a spatial reasoning assistant. You
receive the video frames, a <scene\_context>
block with structured 3D cues, and a <reference\_answer> block. After
understanding the <reference\_answer>, please reason from the video frames
and the structured 3D cues in the <scene\_context> to solve the problem in
your own way and derive the final answer.}

\medskip

\textbf{User Prompt}:

\textit{\{video frames\} \{routed 3D cues\} \{reference
answer\}}

\textit{Question: \{question\}}

\textit{Think step by step and provide your reasoning
inside <think> </think> tags. Then give your final answer inside <answer>
</answer> tags.}
\end{AIbox}

\section{Training Details}

\subsection{Training Dataset}
We construct a 15K-sample training mixture comprising 10K VSI samples, 4K SPAR samples, and 1K MindCube samples. The VSI samples are drawn from VSI-590K~\citep{yang2025cambrian}, while the SPAR and MindCube samples are drawn from the official training splits of their respective benchmarks~\citep{zhang2025spar,wang2025mindcube}. To prevent evaluation contamination, we enforce disjointness between the training mixture and all test sets at both the scene and question levels, ensuring that no scene or question is shared between training and evaluation.

\subsection{Detailed Training Setup}
Table~\ref{tab:training-params} summarizes the main training
hyper-parameters. We implement GPD on top of the
VeRL~\citep{sheng2024hybridflow} framework. All methods and ablations
share the same configuration and differ only in the privilege provided to
the teacher context. Each VSI video is uniformly sampled to 16 frames,
each MindCube sample provides a small set of object views, and SPAR samples
retain their native single-view or multi-view images. During training, every
frame is resized to a pixel budget between $196{,}608$ and $262{,}144$
pixels, which bounds the visual token count per sample. The privileged
teacher re-scoring reuses the same resized frames, so the student and
teacher branches observe identical visual inputs. All experiments are
conducted on two NVIDIA RTX PRO 6000 GPUs (96\,GB each).

\begin{table}[h]
\centering
{\setlength{\tabcolsep}{1mm}%
\begin{tabularx}{\linewidth}{l>{\centering\arraybackslash}X}
\toprule
\textbf{Parameter} & \textbf{Value} \\
\midrule
algorithm.adv\_estimator & grpo \\
rollout\_batch\_size & 128 \\
rollout.n & 8 \\
lr & $1\times10^{-6}$ \\
weight\_decay & $1\times10^{-2}$ \\
optimizer & AdamW (bf16) \\
max\_grad\_norm & 1.0 \\
total\_epochs & 1 \\
rollout.temperature & 1.0 \\
rollout.top\_p & 0.95 \\
max\_prompt\_length & 2048 \\
max\_response\_length & 1024 \\
min\_pixels & 196608 \\
max\_pixels & 262144 \\
privileged\_kl\_coef ($\lambda$) & $3\times10^{-3}$ \\
privileged\_kl\_clip & 1.5 \\
frozen\_base\_kl\_coef ($\beta$) & $1\times10^{-2}$ \\
kl\_penalty\_type & low\_var\_kl \\
correctness\_threshold ($\tau$) & 0.5 \\
entropy\_coeff & 0 \\
n\_gpus & 2 \\
\bottomrule
\end{tabularx}%
}
\caption{Key training hyper-parameters for GPD.}
\label{tab:training-params}
\end{table}

\section{Evaluation Details}

\subsection{Evaluation Dataset}
We evaluate GPD on five benchmark suites used in the main results: VSI-Bench~\citep{yang2025vsibench}, SPAR-Bench~\citep{zhang2025spar}, MindCube~\citep{wang2025mindcube}, MMSI-Bench~\citep{yang2025mmsi}, and ViewSpatial-Bench~\citep{li2025viewspatial}. These benchmarks provide complementary evaluation settings, ranging from object-level geometric relations and video-based spatial understanding to multi-view scene modeling, perspective-taking, and mental simulation.
\begin{itemize}
    \item \textbf{VSI-Bench}: VSI-Bench contains more than 5,000 question-answer pairs from 288 egocentric real-world videos and evaluates spatial perception, memory, and reasoning across diverse indoor environments.

    \item \textbf{SPAR-Bench}: SPAR-Bench contains 7,207 manually verified question-answer pairs spanning 20 spatial tasks under single-view, multi-view, and temporal settings.

    \item \textbf{MindCube}: MindCube comprises 21,154 questions over 3,268 images and evaluates cognitive mapping, perspective-taking, and mental simulation from limited visual observations.

    \item \textbf{MMSI-Bench}: MMSI-Bench contains 1,000 manually curated multiple-choice questions from over 120,000 images, covering grounding, overlap matching, scene reconstruction and transformation, and spatial logic.

    \item \textbf{ViewSpatial-Bench}: ViewSpatial-Bench includes more than 5,700 question-answer pairs from over 1,000 scenes and evaluates spatial localization from both egocentric and allocentric viewpoints.
\end{itemize}

\subsection{Detailed Evaluation Setup}
All evaluations are performed in the student configuration: the model
receives only the RGB observations and the question, and no 3D cue or
reference answer is injected at test time. For VSI-Bench, each video is
uniformly sampled to 16 frames, matching the training setup; the remaining
benchmarks are evaluated with their native single-view or multi-view
images. Evaluation shares the same per-frame pixel budget as training,
with every frame resized to between $196{,}608$ and $262{,}144$ pixels.
We use greedy decoding for all benchmarks. Multiple-choice answers are
parsed from the response by letter matching, and numerical answers are
scored with the threshold-averaged relative accuracy described in the
reward design, so training and evaluation follow a consistent scoring
protocol.

\section{Additional Analysis}

\subsection{Additional Training Dynamics}

\begin{figure}[!t]
\centering
\includegraphics[width=\columnwidth]{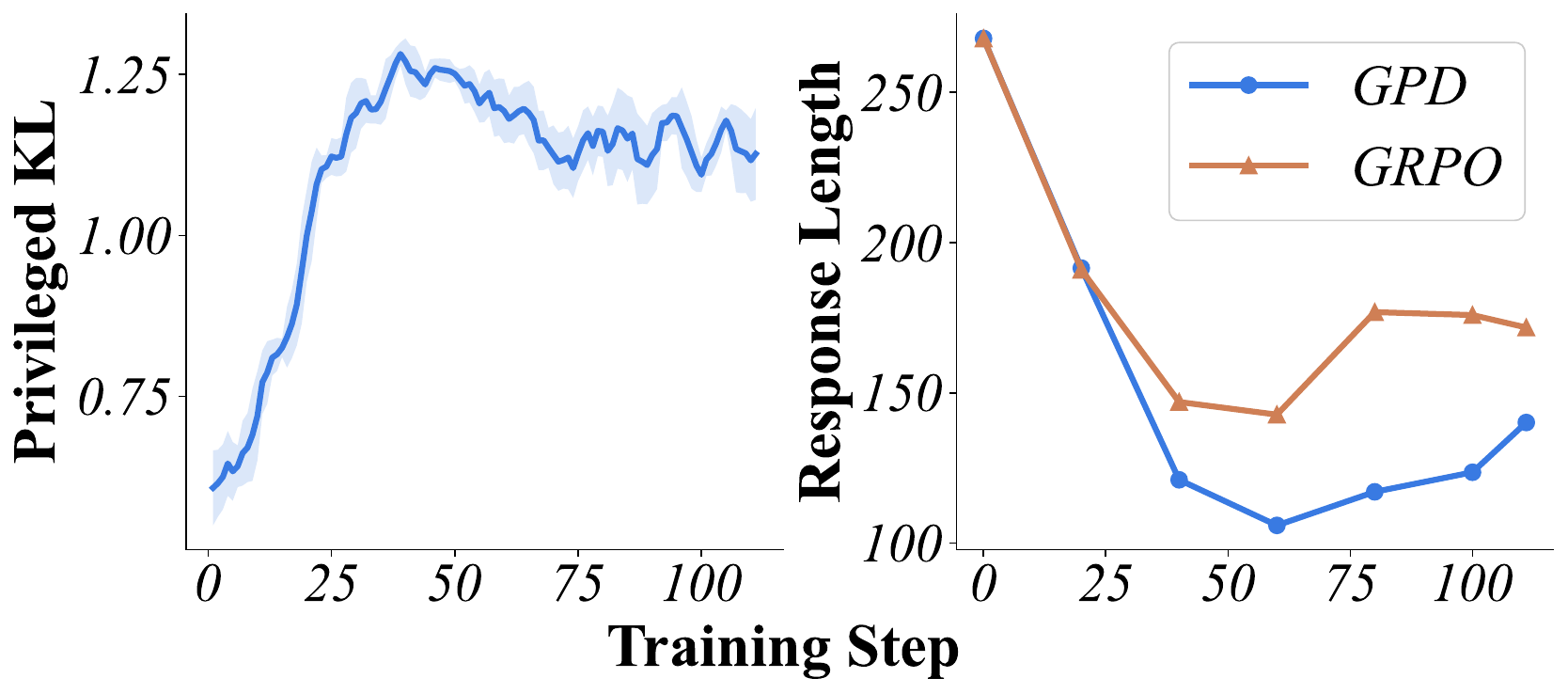}
\caption{Additional training dynamics of GPD and GRPO on Qwen3-VL-4B. Left:
GPD privileged KL. Right: validation response length.}
\label{fig:additional-training-dynamics}
\end{figure}

Figure~\ref{fig:additional-training-dynamics} reveals two complementary trends. First, GPD's privileged KL rises from 0.58 at the first update to 1.30 around step~40, then stabilizes at 1.19 by the end of training, confirming that the geometric teacher signal remains consistently active throughout optimization rather than vanishing as training progresses. Second, GPD produces progressively shorter validation responses after step~40, ending at 140.1 tokens compared to 171.7 for GRPO, an 18.4\% reduction. This response compression suggests that privileged supervision encourages the model to reason more directly toward the correct answer, discarding speculative or redundant intermediate steps that outcome-only supervision cannot prune. Together, the sustained distillation signal and the more concise reasoning chains indicate that GPD's accuracy gains reflect genuinely internalized geometric knowledge rather than longer, more exhaustive search over possible trajectories.

\subsection{Training Cost}

Table~\ref{tab:training-cost} compares GPD and GRPO under matched
training settings. Both runs process approximately 3.35M tokens per step.
GPD increases the average step time from 30.8 to 40.3 minutes, corresponding
to a 30.8\% training-time overhead, while peak allocated memory remains nearly
unchanged. The additional time primarily comes from re-scoring student
trajectories under the privileged teacher context and the frozen-reference
forward pass. Since both branches are used only during training, GPD introduces
no additional inference-time computation or memory cost.

\begin{table}[h]
\centering
\small
\setlength{\tabcolsep}{5pt}
\begin{tabular}{@{}lccc@{}}
\toprule
Method & Time/Step (min) & Total (h) & Peak Mem. (GB) \\
\midrule
GRPO & 30.8 & 57.1 & 64.4 \\
GPD  & 40.3 \textit{(+30.8\%)} & 74.6 & 64.5 \\
\bottomrule
\end{tabular}
\caption{Training cost on Qwen3-VL-4B using two GPUs. Both methods use
111 optimization steps, a global batch size of 128, and a rollout group size
of 8.}
\label{tab:training-cost}
\end{table}

\section{Limitations}
GPD has several limitations. First, privilege quality depends on the accuracy of offline 3D annotation tools and the question-conditioned router; estimation errors in depth or semantic cues propagate directly to the teacher context. Second, the current framework is trained on a limited data mixture of 15K samples, and its scaling behavior under larger training budgets remains unexplored. Third, the privileged teacher may itself produce incorrect token distributions on hard examples, introducing noisy supervision signals that could destabilize training. Finally, while GPD introduces no additional inference cost, privileged re-scoring incurs additional computation during training.

\appendix

\end{document}